\documentclass[
]{ceurart}

\usepackage{listings}
\usepackage{latexsym}
\usepackage{amssymb}
\usepackage{amsmath}
\usepackage{amsthm}
\usepackage{booktabs}
\usepackage{enumitem}
\usepackage{graphicx}
\usepackage{color}

\usepackage{tikz}
\usetikzlibrary{positioning}

\begin{document}

\copyrightyear{2025}
\copyrightclause{Copyright for this paper by its authors.
  Use permitted under Creative Commons License Attribution 4.0
  International (CC BY 4.0).}

\conference{Identity-Aware AI workshop at 28th European Conference on Artificial Intelligence,
  October 25, 2025, Bologna, Italy}

\title{Identity-Aware Large Language Models require Cultural Reasoning}

\author[1]{Alistair Plum}[
orcid=0000-0003-0977-3467,
email=alistair.plum@uni.lu,
]
\cormark[1]
\address[1]{University of Luxembourg, Luxembourg}

\author[1]{Anne-Marie Lutgen}[
orcid=0009-0001-4342-9718,
email=anne-marie.lutgen@uni.lu,
]

\author[1]{Christoph Purschke}[
orcid=0000-0002-9655-2058,
email=christoph.purschke@uni.lu,
]

\author[2]{Achim Rettinger}[
orcid=0000-0003-4950-1167,
email=a.rettinger@uni-trier.de,
]
\address[2]{Trier University, Germany}

\cortext[1]{Corresponding author.}

\begin{abstract}
Large language models have become the latest trend in natural language processing, heavily featuring in the digital tools we use every day. However, their replies often reflect a narrow cultural viewpoint that overlooks the diversity of global users. This missing capability could be referred to as cultural reasoning, which we define here as the capacity of a model to recognise culture-specific knowledge values and social norms, and to adjust its output so that it aligns with the expectations of individual users. Because culture shapes interpretation, emotional resonance, and acceptable behaviour, cultural reasoning is essential for identity-aware AI. When this capacity is limited or absent, models can sustain stereotypes, ignore minority perspectives, erode trust, and perpetuate hate. Recent empirical studies strongly suggest that current models default to Western norms when judging moral dilemmas, interpreting idioms, or offering advice, and that fine-tuning on survey data only partly reduces this tendency. The present evaluation methods mainly report static accuracy scores and thus fail to capture adaptive reasoning in context. Although broader datasets can help, they cannot alone ensure genuine cultural competence. Therefore, we argue that cultural reasoning must be treated as a foundational capability alongside factual accuracy and linguistic coherence. By clarifying the concept and outlining initial directions for its assessment, a foundation is laid for future systems to be able to respond with greater sensitivity to the complex fabric of human culture.
\end{abstract}


\maketitle

\section{Introduction}
\label{sec:intro}
The advent of large language models (LLMs) has brought about considerable advances in the field of artificial intelligence (AI). Language models have allowed many tasks in- and outside of natural language processing (NLP) to be automated. Prominent LLM families such as GPT \cite{Radford2019,OpenAI2024}, LLaMA \cite{Touvron2023}, and Aya \cite{Ustun2024}, which are trained on massive web-scale corpora \cite{Raffel2023}, exemplify the shift towards conversational models that can generate responses to open-domain user questions. AI as a concept has now become more real than ever, and LLMs that ``know'' everything and nothing at the same time are taking us closer.\footnote{We point out that sentences ascribing human capacities to models by using activity verbs or metaphors are not meant in their literal sense, that is, we do not equate model output with human reasoning and speech.}

As we use these LLMs to generate what we desire at ever-increasing rates, concerns about the generated content also arise, often in some way linked to the accurate representation of the diverse cultural identities that shape humanity. More specifically, when we interact with these models, we expect not only correct or reasonable answers, we also expect these models to reason carefully, considering what aspects of our interaction are linked to not facts alone but our identity. Undeniably, our cultural backgrounds stand as a central pillar that constitutes our identity. Because of the sheer diversity of cultures represented on earth, however, it is clear that concerns arise as to the proper or accurate portrayal of these cultures as part of the exchange we have with language models and AI. Language and cultural identity are inherently intertwined \citep{schecter2014language} and therefore the language in which we communicate with AI is central to the understanding of our cultural backgrounds.
These concerns are consistent with evidence that model outputs can encode and amplify social biases unless explicitly addressed in training, data or evaluation design \citep{Blodgett2020BiasSurvey,Bender2021StochasticParrots}.

Recent studies have shown that LLMs trained predominantly on English-language data exhibit biases favouring Western cultural norms, thereby limiting their applicability in non-Western contexts \citep{Joshi2020,Tao2024}. For instance, \citet{Karim2025} show that ostensibly culture-neutral tasks are handled unevenly across cultural contexts, while \citet{Naous2024} demonstrate Western-centric preferences in Arabic settings, with models over-selecting Western entities and frames. The ability to navigate the subtleties of cultural context becomes much harder, however, when models are biased towards Western norms; such bias can lead to outputs that perpetuate stereotypes or marginalise under-represented groups \citep{Hu2025,van2024challenging}, thereby constraining the usefulness of English-dominant LLMs in culturally diverse applications.

Taking all this into consideration, it stands to reason that any efforts that are aimed towards identity-aware AI must be preceded by efforts to improve these issues in LLMs. Culture influences every individual's identity in some way, and this influence is not necessarily bound by country or location. Because of the way culture can influence every individual differently, it is necessary to understand the cultural influences and take them into consideration when interacting with another individual, in effect reasoning while taking culture into account. With AI growing so rapidly in use, we need to ensure that it is inclusive and adaptive, in turn requiring it to be identity-aware and therefore capable of cultural reasoning (CR).  

CR is required when acceptable interpretations and actions depend on local practices rather than universal rules, and where potential bias needs to be processed appropriately. Current evaluations mostly test factual recall or generic safety and therefore do not reveal whether models can recognise culture-specific norms, apply them consistently, and reconcile conflicts across settings. Importantly, bias should not be treated as automatically negative: conceptually, the term bias simply captures the fact that our views on the world as mediated through language put the world in a perspective that reflects our perceptions, norms and beliefs. Where work on raciolinguistic, gender-related or other forms of bias (as the by-product of model training) rightly foregrounds harms and issues with equal reprensentation of cultural diversity, CR clarifies that some forms of “bias” are desirable insofar as they enable simulated perspective-taking, i.e. adopting a culturally situated point of view for appropriate behaviour and interpretation. 

This paper argues that CR is a distinct capability that should be defined, evaluated, and improved with targeted procedures. Beyond model capability claims, this work intersects long-standing concerns about NLP and its social impact and data practices. Early position and survey papers argue that technical progress must be coupled with attention to downstream harms and structural inequities \citep{HovySpruit2016,Blodgett2020BiasSurvey}. Complementary critiques highlight risks from scale and opaque data pipelines \citep{Bender2021StochasticParrots,Paullada2021Data}, and broader taxonomies of language model risks frame why cultural specifics urgently require dedicated evaluation \citep{Weidinger2021Risks,Bommasani2021FoundationModels}.

Efforts to evaluate cultural alignment in LLMs have employed frameworks to assess specific aspects of CR in LLMs. This includes Hofstede's cultural dimensions \cite{Kharchenko2024} and the GLOBE study to assess models' adherence to specific cultural values \cite{Karinshak2024}. However, these approaches often rely on static, survey-based methodologies that may not capture the dynamic and context-dependent nature of culture. They tend to focus on knowledge about cultures rather than the ability to reason within diverse cultural frameworks. In addition, evaluation metrics for LLMs tend to focus primarily on performance benchmarks such as accuracy, fluency, and coherence in tasks like question answering or text summarisation. The evaluation of these cultural capabilities of LLMs on their own does not necessarily show a detailed picture of CR capabilities.

To address the problems and limitations stated above, this paper proposes to define the term \textit{cultural reasoning} in the context of LLMs, as well as the development of a robust, interdisciplinary evaluation framework that transcends traditional NLP metrics. This framework should incorporate methodologies from various domains in humanities research to assess the capacity of LLMs for CR. For instance, evaluating models' responses to culturally sensitive scenarios, their adaptability to context-specific variation in the language use and language choice in multilingual settings, and their understanding of context-specific norms can provide more in-depth insights into their cultural competence.

Furthermore, the case is made for the creation of diverse, representative datasets that encompass a wide range of cultural narratives and perspectives. Initiatives like CulturalBench \cite{Chiu2025} and WorldView-Bench \cite{Mushtaq2025} have made strides in this direction by introducing benchmarks that evaluate LLMs across various cultural contexts. Building upon these efforts, the proposed framework in this paper aims to guide the development and evaluation of AI systems that are not only linguistically proficient but also culturally attuned.

In the following sections, the conceptual underpinnings of CR in AI will be elaborated, a critique of existing evaluation methodologies offered, and an outline of the components of the proposed framework sketched. Through this discourse, we hope that a shift towards more culturally aware AI systems that respect and reflect the rich diversity of human identities can be induced.

\section{Related Work}
\label{sec:related}
The term \textit{cultural reasoning} is not yet an established technical term in AI and appears only sporadically in academic literature. Few papers formally define \textit{culture} or \textit{cultural reasoning}, underscoring the complexity and novelty of the concept \citep{Liu2024, Adilazuarda2024}. Researchers often use adjacent terms like \textit{cultural knowledge}, \textit{cultural awareness}, or \textit{cultural competence} \citep{Chiu2025}. For instance, a recent survey of over 90 papers found that none explicitly define ``culture'' instead probing models with proxy aspects such as values or demographics, and that only certain aspects, such as values, norms, or objectives, have been studied while many remain unexplored \citep{Adilazuarda2024}. 

For the purposes of this paper, we understand culture as defined by \cite{Stalder2018}: culture refers to the dynamic processes through which social meaning is created, negotiated, and materialised in practices, institutions, and lifeworlds. It is not static or merely symbolic, but a contested field shaped by ongoing exchanges, disputes, and power relations that guide both collective and individual activity. Language is part of culture and the primary medium through which culture is expressed, shared, and adapted \cite{kim2003exploring}.

Historically, CR has seen limited use, such as the Cultural-Reasoning Architecture (CARA) system in 2007 \citep{Subrahmanian2007}. CARA was an attempt to model cultural group behaviours and norms for training simulations through a cognitive architecture \citep{Subrahmanian2007}. Today, however, CR is gaining traction within NLP, AI ethics, human-computer interaction (HCI), and cognitive science as AI systems are deployed globally and must navigate diverse cultural contexts. Recent work has begun to work towards how AI can handle cultural differences in knowledge, values, and communication, though there is not necessarily a consensus on definitions and approaches \citep{Liu2025, Adilazuarda2024}. Most research relevant to CR deals with specific sub-areas of this broad challenge. 

\subsection{Cultural Reasoning vs Moral Norms, Bias, and Value Alignment}
Empirical studies suggest that current LLMs have some awareness of cultural differences but limited adaptability in their responses, often related to current cross-lingual architectures of models. For instance, \citet{Kharchenko2024} found that while LLMs can recognise that different countries have differing value orientations (drawing on Hofstede’s cultural dimension theory), they often fail to adjust their advice or reasoning to align with those local values. In other words, a model might know of a cultural norm difference, yet not sufficiently apply that knowledge when generating answers for a user from that culture.  

Closely related to this, \citet{munker2025cultural} found that some models tend to demonstrate clear limitations in the way they capture cross-cultural moral diversity, as they do not sufficiently differentiate between cultural contexts and represent Western perspectives more accurately than non-Western ones.

Similarly, evaluations of moral reasoning across cultures show that LLMs tend to align closely with Western moral frameworks by default, demonstrating a form of cultural myopia in their moral judgments \citep{Ramezani2023}. Without explicit tuning, models largely mirror the norms dominant in their training data, and struggle to accurately reflect the moral or social norms of less-represented cultures. This aligns with broader findings that models often default to majority cultural priors unless representations of norms and trade-offs are made explicit \citep{Sap2020SocialChemistry,Blodgett2020BiasSurvey}.

Beyond values and ethics, other culturally-rooted reasoning tasks reveal significant performance gaps. \citet{Liu2024} investigated multilingual LLMs’ capabilities with proverbs and sayings from different cultures, as these encapsulate cultural wisdom and often require context-specific interpretation. The authors found that state-of-the-art multilingual models know only a limited set of common proverbs and, even when a proverb is memorised, the model may not truly understand its meaning in context \citep{Liu2024}. The models struggled in particular with figurative language and context-based reasoning: for instance, when prompted to interpret or choose the correct continuation of a culturally specific saying, their performance was poor, especially if the question was more complex, such as selecting an incorrect meaning \citep{Liu2024}. Perhaps most striking, a clear culture gap was observed, where models performed worse when reasoning about proverbs translated from languages outside the model’s primary training focus. This suggests that even multilingual LLMs lack robust cross-cultural abstraction and do not seamlessly transfer reasoning skills across cultural contexts. 

Several benchmarks and analyses have reinforced these findings of partial awareness but inadequate adaptation. \citet{Rao2024} introduced NormAD, a dataset of roughly 2,600 short stories from 75 countries, each reflecting local social norms, to test LLMs’ cultural adaptability. Their results showed that models struggle with CR across all levels: whether identifying acceptable behaviour in a story or adapting a continuation to fit a given country’s norms, performance was significantly lower for non-Western contexts \citep{Rao2024}. Even when explicitly provided with the relevant cultural norm as context, the best model, Mistral-7B, achieved only about 82\% accuracy, compared to human performance around 95\%\citep{Rao2024}. Notably, models performed better in judging stories that adhered to common norms than those that violated local norms, hinting at a bias toward assuming normative behaviour, or a general agreeableness bias that impairs detection of culturally deviant situations\citep{Rao2024}. 

In the domain of factual and procedural cultural knowledge, comprehensive evaluations have exposed wide gaps in what LLMs know. \citet{Chiu2025} present CulturalBench, a suite of 1,696 human-written, culturally diverse questions covering 45 global regions (including under-represented ones like Zimbabwe, Bangladesh, and so on) and topics ranging from food and festivals to social etiquette. Human performance on these questions is near 92\% accuracy, yet even top-tier models like GPT-4 struggle. On the hardest version of CulturalBench, state-of-the-art models’ accuracies range roughly between 30\% and 60\% \citep{Chiu2025}. Models often latch onto a single trained-for answer and fail on questions where multiple answers are correct or context-dependent (e.g., ``What utensils do the Chinese usually use'' expects both chopsticks and spoons/forks depending on context, but a model might always answer ``chopstick'' only) \citep{Chiu2025}. According to the authors, model performance is weakest on regions that are less represented in typical training data, such as the Middle East, North Africa, and parts of South America, making it clear that model knowledge is skewed toward cultures prevalent in its training corpus \citep{Chiu2025}. These evaluations echo the pattern seen in norms and values: current LLMs, even when fluent and knowledgeable in a general sense, remain culturally distant and often cannot replicate the breadth of human cultural knowledge or adapt their reasoning to specific cultural settings without additional help. 

Not all findings are entirely pessimistic, though, as some differentiation in outputs across cultures has been observed. For example, when comparing LLMs developed in different cultural environments, there are measurable variations. \citet{Karinshak2024} evaluated Chinese vs. U.S. origin LLMs using the GLOBE cultural values framework and found that each model reflects certain biases of its origin culture’s value system. In their GLOBE benchmark, models showed both similarities and systematic differences in how they prioritise values, suggesting that the cultural context of model development or alignment does influence its behaviour to a degree \citep{Karinshak2024}. However, they also note that extracting these differences requires careful, open-ended analysis and introduce an LLMs-as-a-Jury method for evaluating generation content, since simple QA tests might miss subtle cultural value cues \citep{Karinshak2024}. 

Overall, while LLMs today are not reliably culturally adaptive, research is beginning to show that they possess areas of cultural knowledge and can mimic some cultural differences, but lack a generalised competence to reason as a local across the world’s many cultures. 

\subsection{Culture in Training Data and Representation}
As we have seen, a major reason behind the narrow cultural viewpoint of LLMs is the skewed representation of cultures in their training data. The vast majority of large-scale training corpora for language models are dominated by Western perspectives and English\footnote{It must be noted that there needs to be a distinction between culture and language as they are not interchangeable even though language is part of culture.}, as well as only a few other very high-resource languages. This leads to models that favour Western contexts by default, even when operating in other languages or regions \citep{Naous2024, Joshi2020}. \citet{Joshi2020} quantified the linguistic diversity gap in NLP, showing that a handful of languages, primarily English and a few European and East Asian languages, account for most NLP resources, while the thousands of other languages and by extension, the cultures associated with them, are scarcely represented. Data documentation and dataset development practices further shape which cultural signals are learnable in the first place \citep{Paullada2021Data}. They call into question the language agnostic claims of many models, underlining that current systems inherently prioritise certain cultures unless active efforts are made to include diverse languages \citep{Joshi2020}. In practical terms, this means an LLM is far more likely to have read about Christmas than Diwali, or about New York than Nairobi, yielding an uneven cultural knowledge base. 

Furthermore, by including only small proportions of non-western languages, context-specific varieties and variation is barely included in the training data \cite{Joshi2020, hedderich-etal-2021-survey, blodgett-etal-2016-demographic}. This creates a cultural gap in the communicative style for different settings and favours high prestige varieties that are most likely accounted for in the small amount of training data. In multilingual contexts, this gap widens where language choice for specific situations is a highly complex process. The scarcity of data does not allow for an even representation of different contexts and therefore different languages in the same cultural setting. 

The Western-centric data bias is highly evident in model outputs. \citet{Naous2024} demonstrated that both multilingual and even ostensibly Arabic-focused LLMs showed a strong bias towards Western entities and contexts when tested in Arabic. In their experiments, using their CAMeL benchmark, models frequently produced completions and associations that were more Western-oriented, sometimes even stereotyping non-Western contexts or handling them unfairly \citep{Naous2024}. For example, when generating stories or filling in text, the models struggled to appropriately adapt to specifically Arab cultural settings, often injecting Western assumptions or failing to use culturally appropriate references. Such outcomes are said to be directly traceable to the training data: \citet{Naous2024} analysed common pre-training sources and found them lacking in the richness and versatility needed for culturally aware AI. In fact, they suggest that without significant adjustments, relying on sources like Wikipedia may perpetuate cultural biases since those sources themselves can be skewed or incomplete for certain cultures \citep{Naous2024}. These outcomes are consistent with critiques that large, weakly curated web corpora can entrench existing cultural skews unless counterbalanced \citep{Bender2021StochasticParrots}.

Cultural under-representation in training data also affects fundamental processing at the token level. A follow-up study by \citet{Naous2025} looked at how pre-training data frequencies cause structural biases. They discovered that the frequency-based tokenisation schemes used by LLMs disadvantage less-represented languages and cultural terms. For instance, in Arabic, certain common words or names with multiple meanings  ended up fragmented or poorly encoded by the tokeniser because the model had not seen them in varied contexts often enough \citep{Naous2025}. Moreover, if a language shares script with others, as many non-Arabic languages use Arabic script, tokenisers trained on aggregated text can confuse or conflate culturally distinct terms. As model vocabulary sizes increase, these issues can worsen, leading to higher perplexity and confusion for culturally specific content \citep{Naous2025}. In short, the very way text is ingested by LLMs can reflect cultural biases, as concepts frequent in Western contexts are assigned well-formed embeddings, whereas those frequent in Swahili or Bengali might be less distinct in vector space, hindering the model’s fluency and understanding in those contexts. 

Beyond data imbalance, there are also emergent biases in how models generalise, akin to social identity biases in humans. \citet{Hu2025} examined whether LLMs exhibit in-group vs. out-group bias, a fundamental aspect of cultural psychology. By using prompts such as ``\textit{We are ..., they are ...}'' across many identity group pairs, they found that many base models strongly favour whatever group is described as ``we'' (ingroup favouritism) and often generate derogatory or negative continuations for ``they'' (outgroup) \citep{Hu2025}. This pattern held for various groups and appears to reflect biases present in the training data or human texts. Although instruction-tuned models showed some reduction of this effect, it was still present unless specific bias mitigation fine-tuning was done \citep{Hu2025}. These results imply that an LLM might not just lack knowledge of a culture, but could also unintentionally disrespect it by echoing harmful biases or stereotypes, if those were implicit in the training corpus. From an AI ethics standpoint, this raises concerns about deploying such models globally, as they could reinforce cultural hegemony or prejudice if not carefully corrected. 

Research is actively exploring solutions to these issues of cultural bias and blind spots in data. One straightforward approach is to curate or augment training data with more culturally diverse sources, and to apply fine-tuning to instil cultural knowledge. \citet{Ramezani2023} were able to improve a model’s predictions of various countries’ moral norms by fine-tuning on survey data from those countries. \citet{Hu2025} also showed that careful curation, including balancing training data or filtering out biased content, and additional fine-tuning can substantially reduce the level of in-group/out-group bias exhibited by LLMs. 

Focusing more on the level of the individual, \citet{Zhang2025} clearly demonstrate that current models do not reflect the vast breadth of preferences that individuals have across cultural, political and further dimensions. The authors also compile a dataset to improve model performance in this area.

Another line of research focuses on prompting and multi-agent techniques to inject multiple cultural perspectives. \citet{Mushtaq2025}, for example, argue for a multiplex world-view approach: instead of a single LLM response that might reflect a single dominant perspective, they have multiple LLM agents, each initialised with a different cultural viewpoint, jointly produce an answer. In their experiments, this approach dramatically increased the diversity of perspectives in outputs and improved the overall balance of viewpoints, as measured by an entropy-based metric of perspective distribution \citep{Mushtaq2025}. Such methods highlight that CR may be enhanced not just by feeding more data, but also by architecting interactions, either via prompt engineering or system design that force the model to consider alternatives and lesser-heard viewpoints. 

Finally, the research community is devising more robust benchmarks and evaluation frameworks to track progress in culturally aware AI. In addition to CulturalBench and NormAD mentioned above, others like GIMMICK evaluate vision and language models across many cultural settings to identify where models know tangible cultural facts, such as flags or foods, versus where they fail on intangibles, such as  rituals or values \citep{Schneider2025}. Such evaluations consistently find Western or high-resource culture bias across modalities, but also provide quantitative targets for improvement. The hope is that with clear benchmarks, future models can be trained or adjusted to perform well across all cultures, not just the ones most represented in their training data. Moving forward, the literature points toward a combination of strategies for true CR in AI: better data, meaning more inclusive and representative corpora, better definitions and taxonomies of “culture” to guide what models should learn \citep{Liu2025}, new training or prompting techniques to improve cultural adaptability, and strong evaluation to ensure that as AI becomes culturally competent. 

\subsection{Evaluation of Culture Specifics}
The systematic evaluation of culture specifics also connects to risk taxonomies for language models and the broader analysis of foundation-model externalities \citep{Weidinger2021Risks,Bommasani2021FoundationModels}. A growing body of empirical research and case studies illustrates how LLMs and related AI systems routinely manifest cultural biases and lapses in cultural sensitivity. These include:

\begin{itemize}
  \item \textbf{Western-centric biases in multilingual LLM outputs}: Using the CAMeL dataset focused on Arabic contexts, Naous et al.\ demonstrate that multilingual and monolingual Arabic language models disproportionately favour entities and representations associated with Western culture, leading to inappropriate or stereotyped outputs in Arab cultural settings \citep{Naous2024}.
  
  \item \textbf{Misalignment with culturally specific moral norms}: Tao et al.\ perform a disaggregated evaluation of several LLMs (e.g., GPT-3.5, GPT-4 variants) against nationally representative World Values Survey data. They find that model outputs reflect values more typical of English-speaking and Protestant European societies, rather than those of the countries in question, and while cultural prompting improves alignment for many regions, it fails or even exacerbates bias for others \citep{Tao2024}.
  
  \item \textbf{Propagation of stereotypes and representational harms}: A recent UNESCO backed study revealed that models like GPT-3.5 and Llama 2 engage in regressive gender stereotyping that portrays women predominantly in domestic roles while associating men with career-oriented concepts, as well as displaying homophobic and racial biases \citep{van2024challenging}.
  
  \item \textbf{Salary bias across demographic profiles}: A recent empirical analysis reveals that AI chatbots (e.g., ChatGPT, GPT-4o-mini, Llama 3.1) systematically recommend lower starting salaries to women and ethnic minorities even when qualifications and role descriptions are identical to those of male or white candidates, with differences spanning tens of thousands of dollars \citep{Yamshchikov2025}.
  
  \item \textbf{Subtle dialect-based prejudice}: Reporting on covert forms of racism, researchers found that models such as ChatGPT and Gemini hold biased stereotypes against speakers of African American Vernacular English, perceiving them as less intelligent or employable and resulting in reduced recommendations or harsher judgments \citep{Hofmann2024}.
  
  \item \textbf{Underrepresented cultural groups suffering stereotype bias}: The Indian-BhED dataset reveals that LLMs heavily stereotype Indian-specific axes of identity, such as caste and religion. Models like GPT-2 and GPT-3.5 generated stereotypical outputs in 63–79\% of cases with respect to caste and 69–72\% for religion, underlining their failure to handle Global South contexts sensitively \citep{Khandelwal2023}.
  
  \item \textbf{Stereotype propagation across multiple languages}: The SHADES dataset, created under the BigScience initiative, allows systematic measurement of stereotypes in multiple languages. SHADES reveals that models often replicate harmful stereotypes beyond English, and even fabricate pseudo-scientific justifications for them, therefore extending cultural harm across linguistic boundaries \citep{Rogers2025}.
\end{itemize}

These documented issues with model output underscore a critical problem: current LLMs lack genuine cultural understanding and, without disciplined intervention, amplify cultural bias in ways that are often subtle, damaging, and widespread.

\section{Defining Cultural Reasoning}
\label{sec:def}
CR in AI is an emerging field that is investigating how AI systems understand and adapt to the world’s diverse social and cultural norms. Current LLMs exhibit a degree of cultural knowledge and can mimic some differences, but they largely remain biased towards their dominant cultural context of their training data and struggle with truly adapting to unfamiliar cultural scenarios. The research so far, which spans moral reasoning, value alignment, idiomatic understanding, and bias analysis, shows a clear picture of the challenges. It also lays important groundwork, by identifying specific shortcomings, ranging from incorrect moral norm predictions to proverb misinterpretations, these studies guide the development of methods to make AI more culturally aware. Placing CR within these established lines of critique (bias surveys, data-centric risks, and LM risk taxonomies) clarifies both why the capability matters and how progress should be measured \citep{HovySpruit2016,Blodgett2020BiasSurvey,Paullada2021Data,Weidinger2021Risks}.

In this paper, \emph{cultural reasoning} denotes the capacity to select and justify interpretations or actions that are contingent on culture-specific norms, conventions, and procedures, given a locale and context. It differs from cultural \emph{facts}, style or register control (lexical or politeness choices), moral value judgement per se, and generic bias mitigation, although it can include any of these. We work with the definition of culture already mentioned in Section \ref{sec:related}. In addition, while some recent systems expose intermediate ``reasoning'' tokens \citep{OpenAI2025}, we use the term in a broader, model-agnostic sense, and more in line with its traditional meaning.

We treat CR as distinct from translation/localisation alone and from demographic targeting. Producing the right language variety or tone is necessary but not sufficient; the system must use culture-specific premises to arrive at, and explain, its choice in a manner that is adequate for the context. Related strands on normative and social common-sense reasoning provide scaffolding for expressing such culture-specific justifications \citep{Sap2020SocialChemistry}.

We also place CR carefully in relation to bias and de-biasing efforts in research. While we have shown at length the problematic tendencies that arise with bias in training data, we do not regard de-biasing training data as part of CR. Much rather, we consider the detection and careful treatment of bias in these situations as part of CR.

In operational terms, a model response is taken to \emph{exhibit cultural reasoning} if it satisfies one or more of the following:
\begin{enumerate}[label=(\alph*), leftmargin=2.2em]
  \item \textbf{Context-appropriate application:} selects procedures or norms that vary by locale (e.g., administrative steps, forms of address) and applies them correctly to the described situation.
  \item \textbf{Justified adaptation:} provides a short rationale that references relevant norms, practices, or constraints for the specified cultural frame.
  \item \textbf{Conflict reconciliation:} when multiple cultural frames are salient (e.g., cross-border or diasporic settings), reconciles them explicitly, with prompts such as ``do X because Y takes precedence in setting S''.
  \item \textbf{Sensitivity to intra-cultural variation:} acknowledges plurality (majority/minority practices; regional/age variation) with calibrated uncertainty or requests for disambiguation when appropriate.
  \item \textbf{Pragmatic interpretation:} interprets culturally bound figurative language, implicatures, or rituals in context rather than giving literal or default-global readings.
  \item \textbf{Consistency under paraphrase/contrast:} makes stable choices across rephrasings and structured contrast sets tied to the same cultural premise.
\end{enumerate}


\section{Analysing and Evaluating Cultural Reasoning in LLMs}
\label{sec:framework}
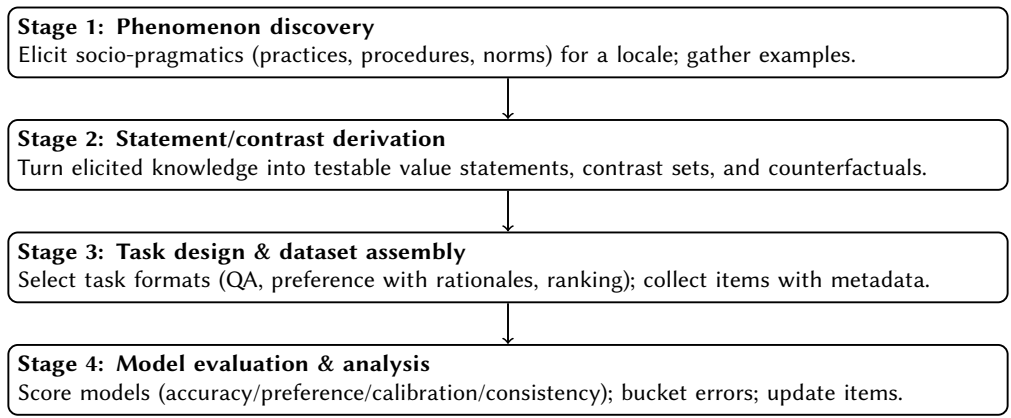
\begin{figure}[ht!]
\centering
\scalebox{0.9}{%
\begin{tikzpicture}[
  font=\small,
  node distance=6mm,
  stage/.style={rectangle, rounded corners, draw, thick, align=left, inner sep=4pt, text width=0.9\linewidth}
]
\node[stage] (s1) {\textbf{Stage 1: Phenomenon discovery}\\
Elicit socio-pragmatics (practices, procedures, norms)  for a locale; gather examples. };

\node[stage, below=of s1] (s2) {\textbf{Stage 2: Statement/contrast derivation}\\
Turn elicited knowledge into testable value statements, contrast sets, and counterfactuals.};

\node[stage, below=of s2] (s3) {\textbf{Stage 3: Task design \& dataset assembly}\\
Select task formats (QA, preference with rationales, ranking); collect items with metadata.};

\node[stage, below=of s3] (s4) {\textbf{Stage 4: Model evaluation \& analysis}\\
Score models (accuracy/preference/calibration/consistency); bucket errors; update items.};

\draw[->, thick] (s1) -- (s2);
\draw[->, thick] (s2) -- (s3);
\draw[->, thick] (s3) -- (s4);
\end{tikzpicture}%
}
\caption{Four-stage cultural-reasoning evaluation pipeline.}
\label{fig:pipeline}
\end{figure}

To better analyse and evaluate CR in LLMs, we propose the following methodology, which aims to systematically elicit, validate, and integrate culturally specific knowledge into large language models in order to improve their ability to engage in culturally sensitive reasoning. The process is organised into several sequential stages (see Figure \ref{fig:pipeline}), each building on the outcomes of the previous one.

\subsection{Identifying Domains of Cultural Variation}
The first stage involves identifying a set of domains or areas of life in which cultural values, norms, and procedures differ significantly across societies. These may include, for example, family structure, workplace hierarchy, gift-giving customs, conflict resolution strategies, and moral priorities. The selection will draw on established cross-cultural psychology frameworks (e.g., Hofstede's dimensions, Schwartz's value theory) as well as ethnographic and sociolinguistic literature. These domains will serve as thematic anchors for all subsequent data collection and evaluation.

\subsection{Eliciting and Evaluating Cultural Descriptions}
For each domain, prompts will be designed to elicit descriptive accounts of the relevant norms or values in multiple languages for multiple target countries, forming an initial matrix of culture–language combinations. A further aspect that we will consider in this matrix is the importance of specific languages and varieties for specific domains in multilingual cultural settings. 

Taking Luxembourg as an example \cite{EhrhartFehlen2011}, a country with three official languages, Luxembourgish (the national language), French (the legislative one) and German, and with a deep cultural history of language contact from the neighbouring regions, Germany, France and Belgium. One domain, where the language choice is very clear is the legal one, as French is institutionally the only language allowed for. However, since the multilingual situation is very complex in Luxembourg, the choice of language in a communicative situation is complex and relies on cultural knowledge and the language of the communicative recipient. This situation is even more complicated considering the high percentage of non-Luxembourgers living in the country and the high number of workers crossing the border every day from France, Belgium and Germany. As this is only one example for a multilingual society from many, it is clear that language plays a big part in the cultural setting. Table \ref{tab:domain_matrix} shows an overview of what this setup could look like conceptually for examples from the Greater Region of Luxembourg, Wallonia, Lorraine, Rhineland-Palatinate, and Saarland (Greater Region of SaarLorLux).

Further, the role of English will be carefully considered, both as a native cultural context and as a cross-cultural medium, to capture potential differences in meaning when norms are expressed in English rather than the local language. The resulting descriptions will then be evaluated by human annotators familiar with the relevant cultural contexts, who will indicate whether each description is accurate (\textit{is true}) or inaccurate (\textit{is not true}). This evaluation step ensures that only reliable and culturally authentic descriptions progress to the next phase.

\subsection{Deriving and Assessing Value Statements}
From the verified descriptions, concise value statements will be formulated to capture the underlying principles, priorities, or normative expectations reflected in the data. These statements should be distilled from the descriptions in a way that retains cultural specificity while also being general enough to apply across related contexts. Once drafted, the value statements will undergo a second human evaluation in which annotators judge three dimensions: Whether the statement is true for the target culture, the importance of the value within that culture and the specificity of the statement, making sure it is neither too vague to be meaningful nor so narrow as to lose relevance. This step filters the statements to retain only those that are both culturally valid and useful for downstream modelling.

\begin{table}[!ht]
\centering
\scriptsize
\begin{tabular}{p{3cm} p{2cm} p{4cm} p{4cm}}
\toprule
\textbf{Domain} & \textbf{Country–Lang.} & \textbf{Example Prompts} & \textbf{Importance of Lang./Var. } \\
\midrule
Family \& Child-rearing & LU–lb/fr/de/pt/en & How do parents talk to young children? / What values are taught at home? & \textit{lb} dominant in families; \textit{fr} in early education; \textit{pt} central in diaspora homes. \\ \\ \hline

Education \& Schooling & DE–de/en & What makes a good teacher? / How do students address authority? & \textit{de} as instructional norm; \textit{en} growing in schools. \\ \\ \hline

Work \& Professional Life & FR–fr/en & How should one behave with a boss? / How formal is workplace communication? & \textit{fr} used in formal settings; \textit{en} in international work. \\ \\ \hline

Public Services \& Administration & LU–fr/lb/de & How are citizens addressed in official communication? & \textit{fr} dominates bureaucracy; \textit{lb} adds local tone; \textit{de} used cross-border. \\ \\ \hline

Media \& Public Discourse & Greater Region \break –lb/fr/de & How are political opinions expressed in local media? & Mix varies regionally; choice of language indexes stance or identity. \\ \\ \hline

Everyday Politeness \& Interaction & FR–fr & How do people greet or thank each other? & \textit{fr} sets norms; regional varieties convey familiarity or solidarity. \\ \\ 

\bottomrule
\end{tabular}
\caption{Matrix of domains, country–language combinations, example prompts, and cultural relevance.}
\label{tab:domain_matrix}
\end{table}


\subsection{Fine-Tuning and Post-Evaluation}
The subset of high-quality value statements will be used to fine-tune a language model with the goal of improving its ability to recognise, articulate, and apply culturally relevant values in its outputs. After fine-tuning, the original prompting procedure for cultural descriptions will be repeated, and the new outputs will be evaluated using the same \textit{is true} / \textit{is not true} criterion as before. Comparing pre- and post-fine-tuning results will indicate whether the inclusion of validated value statements has led to measurable improvements in the cultural accuracy, contextual relevance, and nuance of the model’s responses.


\section{Discussion}
The synthesis of prior work and our empirical considerations clarify why \textit{cultural reasoning} should be treated as a cornerstone capability for identity-aware AI. While moral norms are important, culture also includes pragmatic conventions, historical narratives, communicative styles, and procedural knowledge, as has been argued previously \citep{Zhou2025a,Adilazuarda2024}. This distinction matters, because even when certain moral norms appear to generalise across cultures \citep{Ramezani2023}, their interpretation and salience are culturally situated and interact with other values \citep{Tao2024,Kharchenko2024}.

While we have shown that models and data are often Western-centric, and that areas outside this boundary are disadvantaged, the need for CR lies also within this boundary. For example, the Greater Region of SaarLorLux illustrates the practical complexity of CR. Here, norms and languages mix across borders, producing hybrid identities in which people switch between, or merge, cultural frames. The Greater Region encompasses regions from four different countries and even more language varieties than only German, French and Luxembourgish are part of this area. Moreover, the multilingual situation is central to this region and even differs in each country.

The importance of including the sociolinguistic situation in the thought process is implied through this example. As language and culture are deeply intertwined and form identities, this holds even more importance in multilingual settings. As the choice of language or variety in specific situations is a complex process that is highly routinised in the society itself. However, for AI this is part of the cultural knowledge that it needs to be aware of. Effective AI must recognise distinct norms and, when appropriate, reconcile them through a process that applies relevant cultural information in new mixed contexts.

CR therefore goes beyond cultural awareness. Awareness recognises differences, whereas reasoning applies, adapts, and, when necessary, combines norms to guide behaviour or generation, following \citet{Liu2025,Chiu2025}. Eliciting this capability in LLMs is difficult, because a model must ``behave'' as a person from one culture, or a blend of cultures, would behave, incorporating implicit values, communicative preferences, and context-specific interpretations \citep{Park2023,Naous2025}.

Cultural bias is related but distinct. Bias reflects the sum of learned perspectives on a given subject and often arises from  pre-training data, as shown by a lot of current research \citep{Naous2024,Joshi2020}. Such bias can undermine CR by triggering defaults to overrepresented norms even when another frame is more appropriate \citep{Hu2025,van2024challenging}. Recent benchmarks, including NORMAD by \citet{Rao2024}, CulturalBench by \citet{Chiu2025}, and SeaEval by \citet{Wang2024}, report frequent misapplication of norms in under-represented contexts and suggest the need for targeted interventions.

It is tempting to assume that modern models can adopt a cultural perspective on demand, for example when asked to answer as a person from a less well-represented country. Models often simulate well-represented cultures \citep{Mushtaq2025,Liu2024}, yet success depends on the extent and quality of cultural signals in training. This dependence raises questions about how much knowledge is sufficient and whether models can integrate multiple influences, including whether an LLM could emulate a user shaped equally by Luxembourgish and French norms. Progress requires new evaluation methods \citep{Karinshak2024,Li2024} and clearer theory on what constitutes sufficient cultural competence.

Cross-lingual transfer should be distinguished from cross-cultural transfer. Multilingual models often transfer linguistic form effectively \citep{Lim2025,Yong2024}, yet nuanced CR is far harder. Subtle cues, including indirectness in politeness strategies or the moral weight of certain actions, do not reliably survive translation \citep{Karim2025,Sinelnik2024}. Identity-aware AI therefore cannot rely on multilingual capacity alone and must model culture-specific reasoning directly.

In sum, CR aligns system outputs with users’ cultural contexts in ways that exceed surface-level adaptation. For identity-aware AI, this capability is necessary for equitable and effective interaction. Achieving it will require curated cultural knowledge, mitigation of bias at data and model levels, and evaluation frameworks that reflect the complexity of real-world, often mixed, cultural settings.


\section{Conclusion}
This paper presented a survey of recent work at the intersection of NLP, HCI, and computational social science to situate CR within existing concepts and evaluations. Concretely, we contrasted adjacent notions (cultural knowledge, awareness, alignment) with CR and synthesised evidence from benchmarks and studies showing that contemporary LLMs often default to Western-centric norms and struggle to adapt to under-represented cultural contexts. This review established both the problem space and the measurement gap that motivates our approach.

We proposed an operational definition of CR as the capability to recognise, apply, and combine culturally grounded norms, values, and procedures to guide model behaviour. Methodologically, we distinguished CR from mere awareness by requiring perspective-taking and norm-application, including in mixed or hybrid identities (e.g., users in the Greater Region). The definition is harmonised with ML usage of “reasoning” while remaining agnostic to implementation, thereby supporting evaluation across architectures and training regimes.

Moreover, we outlined a concrete, data-to-evaluation pipeline. The pipeline includes the identification of domains of cultural variation, as well as the elicitation of multilingual country-specific descriptions via prompting. Going beyond these steps, the proposed pipeline includes human validation  as well as fine-tuning steps to re-evaluate and improve model performance in terms of CR. This design deliberately goes beyond knowledge, supports mixed-culture prompts, and provides pre and post metrics tied to human judgements.

We argued that CR is necessary for identity-aware AI because moral norms alone are insufficient and because real users often inhabit blended cultural contexts. We separated cultural bias (distributional skew that pushes models toward dominant frames) from CR (the ability to apply the right frame), while noting that bias directly impairs CR. We also clarified why multilingual transfer does not guarantee cultural transfer, as language form can move across tongues more readily than nuanced cultural priors such as politeness strategies, norm salience, or situated moral trade-offs.

Overall, this paper should provide a practical path to measure and improve cultural reasoning in LLMs. The approach treats CR as testable behaviour grounded in validated cultural descriptions and value statements, enables assessment in mixed-identity scenarios, and yields actionable signals for data curation and fine-tuning. This positions identity-aware AI not as a vague aspiration but as an achievable target: models that reliably adapt to users’ cultural contexts, while preserving clarity, safety, and utility.

\begin{acknowledgments}
This paper was written as part of the UniGR-Guest Professorship project ``Cultural Reasoning in AI: Examining Language Models in the Context of the Greater Region'', awarded and funded by the University of the Greater Region. We thank UniGR for its support.
\end{acknowledgments}

\section*{Declaration on Generative AI}
During the preparation of this work, the authors used GPT-5 and LanguageTool for grammar, style and spell checks. After using these tools, the authors reviewed and edited the content as needed and take full responsibility for the publication’s content.

\bibliography{sample-ceur}

\end{document}